\documentclass[runningheads]{llncs}
\usepackage[T1]{fontenc}
\usepackage{graphicx,verbatim}
\usepackage{amsmath}
\usepackage{multirow}
\usepackage{amssymb}
\usepackage{graphicx}
\usepackage{xcolor}
\usepackage{hyperref}
\hypersetup{
    citecolor=blue,
    colorlinks=true,
    linkcolor=blue,
    filecolor=magenta,      
    urlcolor=cyan,
}
\usepackage{marvosym}
\begin{document}
\title{CPS$^4$: Class Prompt driven Semi-Supervised Spine Segmentation with Class-specific Consistency Constraint}
%

\authorrunning{Qingtao Pan et al.}
\author{Qingtao Pan\textsuperscript{1,3}, Hongzan Sun\textsuperscript{(\Letter)2}, Bing Ji\textsuperscript{(\Letter)1}, and Shuo Li\textsuperscript{3,4}}  
\institute{
School of Control Science and Engineering, Shandong University, Jinan, Shandong 250061, China\\
\email{b.ji@sdu.edu.cn}
\and
Department of Nuclear Medicine, Shengjing Hospital of China Medical University, Shenyang, Liaoning 110004, China\\
\email{sunhongzan@126.com}
\and
Department of Computer and Data Science, Case Western Reserve University, Cleveland, OH 44106, USA
\and
Department of Biomedical Engineering, Case Western Reserve University, Cleveland, OH 44106, USA
}

\maketitle              
\begin{abstract}
Vision-Language Model (VLM) has great potential to enhance the quality of pseudo labels in semi-supervised spine segmentation by leveraging textual class prompts to generate segmentation map, but no one has studied it yet. Although promising, it lacks explicit constraints to ensure consistency between spine class prompts and spine unit region, resulting in unsatisfactory performance in multi-class segmentation map generation. In this paper, we propose CPS$^4$, the first text-guided semi-supervised spine segmentation network using class prompts to enhance the quality of spine pseudo labels. Specifically, CPS$^4$ is implemented through two training stages. (i) Class-specific consistency constrained VLM pretraining stage: we propose token- and pixel-level attention loss to optimize the consistency between class prompts and spine units, forcing the textual class prompt to be closely coupled with the target spine unit in the semantic space. (ii) Class Prompt driven semi-supervised spine segmentation stage: using the pretrained vision–text encoder, we derive each class-specific binary segmentation map for the unlabeled spine image and integrate them into an unified multi-class segmentation map, improving the quality of the spine pseudo label generated by the semi-supervised spine segmentation network. Experimental results show that our CPS$^4$ achieves superior spine segmentation performance with Dice of 80.44\%, only using 5\% labeled data on the public spine segmentation dataset, surpassing popular semi-supervised learning and VLM methods. Our code will be available.


\end{abstract}

\begin{figure*}[t]
\centering
\includegraphics[width=\linewidth]{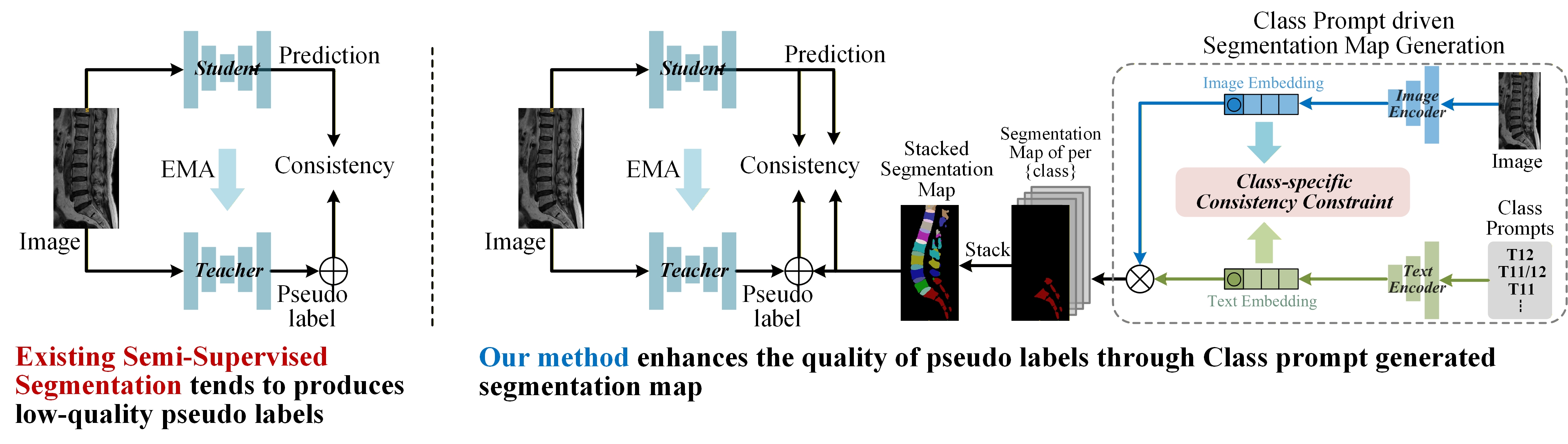}
\caption{Our proposed CPS$^4$ generates class prompt guided segmentation map, enhancing the quality of pseudo labels in semi-supervised spine segmentation.}
\label{fig1}
\end{figure*}

\section{Introduction}
Automatic spine segmentation, i.e., multi-class segmentation of the vertebral bodies (VBs) and intervertebral discs (IVDs) in spine magnetic resonance (MR) images, plays a significant role in various orthopedic applications, including spinal disease diagnosis \cite{ref1}, surgical treatment planning \cite{ref2}, and spine pathology identification \cite{ref1}. Medical image segmentation methods have made significant progress relying on a significant amount of labeled data \cite{ref3}. However, labeling pixel-level annotations is laborious and demands expertise, particularly in spine images with multiple categories. Unlabeled data, on the contrary, are cheap and relatively easy to acquire. Therefore, semi-supervised learning (SSL) methods have the capability to mitigate the label scarcity problem by using a limited amount labeled data and any quantity of unlabeled data \cite{ref4}.

Existing SSL methods are usually based on consistency regularization \cite{ref5} and pseudo labeling \cite{ref6}. Conducting consistency regularization and pseudo labeling between the sub-networks (Fig. \ref{fig1} (\textit{left})), have achieved favorable performance in semi-supervised segmentation \cite{ref7,ref8}. However, these approaches may suffer from low-quality pseudo labels \cite{ref21,ref15b}, which introduce noisy supervision and mislead the learning process. Such inaccurate pseudo labels can cause the model to focus on incorrect features, ultimately preventing effective exploitation of unlabeled data. The question that comes to mind is: \textit{how to effectively improve the quality of pseudo labels for SSL}.

Vision-Language Models (VLMs) \cite{ref9,ref10,ref11,ref12,ref24} have great potential to enhance the quality of pseudo labels. It learns correlations between images and text to understand the semantic information in unlabeled images through textual descriptions, and can generate class prompt guided segmentation map to improve the quality of pseudo-labels \cite{ref22}. However, current VLMs are hard to directly apply to semi-supervised spine segmentation. This is because they lack explicit constraints to ensure consistency between class prompts and the corresponding image objects \cite{ref13}, which weakens class-specific semantic binding. Such insufficient supervision introduces cross-class interference, leading to poor performance in class prompt guided segmentation map generation.

\begin{figure*}[t]
\centering
\includegraphics[width=\linewidth]{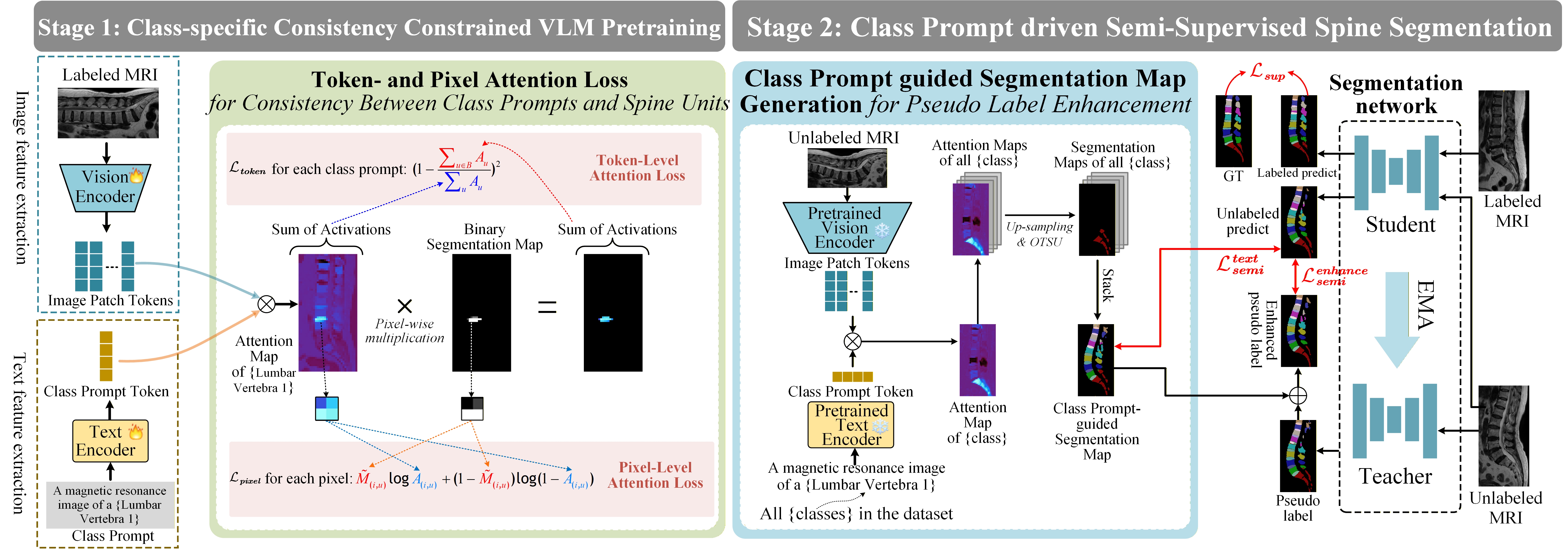}
\caption{Overview of our CPS$^4$. In stage 1, we perform class-specific consistency constrained VLM pretraining to train the vision and text encoder, where token- and pixel-level attention losses are proposed to boost the consistency between class prompts and spine units. Note: In stage 1, we only show the calculation process of \{Lumbar Vertebra 1\} class, for other classes in the MRI, the calculation process is consistent. In stage 2, we conduct class prompt–driven semi-supervised segmentation to train the segmentation network, where the pseudo label is enhance through class prompt generated segmentation map.}
\label{fig2}
\end{figure*}

In this paper, we propose CPS$^4$ (Fig. \ref{fig1} (\textit{right})), a Class Prompt–driven Semi-Supervised Spine Segmentation network that leverages class prompt information to enhance pseudo-label quality. CPS$^4$ is achieved through two-stage training strategy. (i) In the class-specific consistency constrained VLM pretraining stage, token- and pixel-level attention losses are introduced to enforce the consistency between class prompts and spine units, encouraging tight semantic coupling between each class prompt and its target spine unit. (ii) In the class prompt–driven semi-supervised segmentation stage, leveraging the pretrained vision–text encoder, we generate each class-specific binary segmentation map for the unlabeled spine image and aggregate them into a unified multi-class segmentation map, which is used to improve the quality of spine pseudo labels produced by the semi-supervised segmentation network. Our contributions include:

(1) This is the first work that introduces textual class prompts to the semi-supervised spine segmentation. The proposed CPS$^4$ enhances the quality of spine pseudo labels for effective semi-supervised spine segmentation.

(2) A novel class-specific consistency constraint is proposed. We design token- and pixel-level attention losses to strengthen the consistency between class prompts and spine units, imposing explicit constraints for each spine unit.

(3) The experimental results on the spine MRI dataset demonstrate that our method achieves state-of-the-art across different labeled ratios.

\section{Methodology}
Our CPS$^4$ contains two stages. Stage 1: Class-specific consistency constrained VLM Pretraining (Fig. \ref{fig2} (\textit{left})). We propose token- and pixel-level attention losses to enforce the consistency between class prompts and spine units. Stage 2: Class prompt driven semi-supervised spine segmentation (Fig. \ref{fig2} (\textit{right})). We use the pretrained vision–text encoders to generate each class-specific binary segmentation map for the unlabeled spine image and aggregate them into a unified multi-class segmentation map, improving spine pseudo labels.

\subsection{Class-specific Consistency Constrained VLM Pretraining}
Existing VLMs (such as CLIP and MedCLIP) perform cross-modal alignment at the image level, which causes distinct class prompts to fail to focus on its corresponding objects appeared in the image, resulting in poor capabilities in generating class prompt guided segmentation map. To alleviate this issue, token- and pixel-level attention losses are proposed to enable class-specific consistency constraint for VLM Pretraining, guiding each class prompt to align with its corresponding spine unit.

\vspace{4pt}
\noindent
\subsubsection{Token-Level Attention Loss.}
For each class prompt $i$, we extract its class prompt token $F_{T,i}$ using text encoder and acquire the binary segmentation map $M_{i}$ from its respective image. For the spine image, we extract its patch tokens $F_{I}$ using vision encoder. Subsequently, we take $F_{T,i}$ to compute the similarity score with every image patch token $F_{I,j}$, generating the coarse attention map $A_{i}$. The resolution of $M_{i}$ is downscaled to match its corresponding $A_{i}$ with bilinear interpolation, followed by binarization of all values to form $\tilde{M_i}$.

The token-level attention loss $\mathcal{L}_{\mathrm{token}}$ aggregates activations of attention map toward predicted spine unit $\mathcal{B}_i=\{u\in\tilde{M_i}|u=1\}$. $\mathcal{L}_{\mathrm{token}}$ is defined as follows,
\begin{equation}
\mathcal{L}_{\mathrm{token}}=\frac{1}{N}\sum_{i}^{N}\left(1-\frac{\sum_{u\in\mathcal{B}_i}A_{(i,u)}}{\sum_{u}A_{(i,u)}}\right)^2
\end{equation}
where $A_{(i,u)}$ represents the scalar attention activation at a spatial location $u$ of $A_{i}$ formed by the class prompt token $F_{T,i}$ and image patch tokens $F_{I}$. $N$ is the number of classes of the corresponding spine image.

\subsubsection{Pixel-Level Attention Loss.}
Although $\mathcal{L}_{\mathrm{token}}$ substantially aggregates activations of attention maps toward the target spine unit, a side effect of this aggregation is that the VLM tends to overly aggregate its activations of the attention map into certain subregions of its target spine unit. To overcome this problem, we use pixel-level attention loss $\mathcal{L}_{\mathrm{pixel}}$ to counteract. Formally, for a attention map $A_{i}$, we add pixel-level cross-entropy objective $\mathcal{L}_{\mathrm{pixel}}$, which is defined as,
\begin{equation}
\mathcal{L}_{\mathrm{pixel}} =-\frac{1}{N}\sum_i^{N}[\tilde{M}_{(i,u)}\log A_{(i,u)}+\left(1-\tilde{M}_{(i,u)}\right)\log\left(1-A_{(i,u)}\right)]
\end{equation}

We utilize $\mathcal{L}_{\mathrm{token}}$ and $\mathcal{L}_{\mathrm{pixel}}$ to optimize vision and text encoders, i.e., the optimization loss of the pretraining stage is $\frac{1}{2} \left( \mathcal{L}_{\mathrm{token}} + \mathcal{L}_{\mathrm{pixel}} \right)$

\subsection{Class Prompt driven Semi-Supervised Spine Segmentation}
\subsubsection{Class Prompt-guided Segmentation Map Generation.} 
The spine image patch tokens are merged with text prompts for each class, generating segmentation map corresponding to each class prompt. Specifically, leveraging the vision and text encoders pretrained in stage 1, the attention map per class is generated by dot product between the image patch tokens $F_{I}$ encoded by the vision encoder and the class prompt token $F_{T,i}$ encoded by the text encoder. Then, Up-sampling is applied to refine attention map and OTSU automatic threshold method is used to convert the attention map into a binary segmentation map $y^{text}_i$. Finally, the segmentation maps of all classes are stacked, forming the final class prompt-guided segmentation map $y^{text}$.
\begin{equation}
y^{text}_i = \operatorname{OTSU}(\operatorname{Up}(F_{I}\otimes F_{T,i}^{\top} )) \quad y^{text} = \operatorname{stack}_{i\in N}(y^{text}_i),
\end{equation}
where $\operatorname{Up}(\cdot)$ is the Up-sampling operation, $\operatorname{OTSU} (\cdot)$ is the OTSU operation, and $\operatorname{stack}(\cdot)$ integrates each $y^{text}_i$.

\subsubsection{Pseudo Label Generation.} 
Mean Teacher (MT) \cite{ref5} is used for the baseline of semi-supervised segmentation network. It includes two sub-segmentation networks, i.e., student and teacher. The teacher network $f_{\theta_t}$ is updated by the student network $f_{\theta_s}$ via the Exponential Moving Average (EMA). $f_{\theta_s}$ predicts the labeled image $x_l$ and the unlabeled image $x_u$. $f_{\theta_t}$ generates the pseudo-label $y_u^{t}$ of the unlabeled image.
\begin{equation}
y_l=f_{\theta_s}(x_l),\quad y_u^s=f_{\theta_s}(x_u),\quad y_u^t=f_{\theta_t}(x_u),
\end{equation}
where $y_l$ is the labeled prediction, and $y_u^s$ is the unlabeled prediction.

Both the supervised loss $\mathcal{L}_{sup}$ and semi-supervised loss $\mathcal{L}_{semi}$ are utilized to jointly optimize the segmentation network. For labeled images $x_l$, $\mathcal{L}_{sup}$ is calculated between $y_l$ and ground truth $y_{g}$. For unlabeled images $x_u$, $\mathcal{L}_{semi}$ is achieved by two semi-supervised objectives $\mathcal{L}_{semi}^{text}$ and $\mathcal{L}_{semi}^{enhance}$, where $\mathcal{L}_{semi}^{text}$ is calculated between $y_u^{s}$ and $y^{text}$, and $\mathcal{L}_{semi}^{enhance}$ is computed between $y_u^{s}$ and the enhanced pseudo label generated by integrating pseudo-label $y_u^{t}$ and $y^{text}$.
\begin{equation}
\mathcal{L}_{sup}=-\frac{1}{N_l}\frac{1}{HW}\sum_{n=1}^{N_l}\sum_{m=1}^{HW}\ell_{ce}(y_{l_{i,j}},y_{g_{i,j}}),
\end{equation}
\begin{equation}
\mathcal{L}_{semi}^{text}=-\frac{1}{N_u}\frac{1}{HW}\sum_{n=1}^{N_u}\sum_{m=1}^{HW}\ell_{ce}(y_{u_{n,m}}^{s},y_{_{n,m}}^{text}),
\end{equation}
\begin{equation}
\mathcal{L}_{semi}^{enhance}=-\frac{1}{N_u}\frac{1}{HW}\sum_{n=1}^{N_u}\sum_{m=1}^{HW}\ell_{ce}(y_{u_{n,m}}^{s},(y_{u_{n,m}}^{t}+y_{_{n,m}}^{text})),
\end{equation}
where $\ell(\cdot)$ is the cross-entropy loss. $(n,m)$ represents the $m$-th pixel in $n$-th sample. $N_u$ is the number unlabeled images and $N_l$ is the number of labeled image. $W$ and $H$ represent the width and height of an image. Therefore, the total optimization objective of stage 2 is $\frac{1}{2} \left( \mathcal{L}_{semi}^{text} + \mathcal{L}_{semi}^{enhance} \right) + \mathcal{L}_{sup}$

\begin{table*}[h]
  \centering
  \caption{The comparative experiments on MRSpineSeg dataset demonstrate the strong semi-supervised spine segmentation capability of our method. The mean Dice and mIoU values are reported and the best results are highlighted in bold.}
  \resizebox{0.95\textwidth}{!}{
    \begin{tabular}{l|c|c|c|c|c|c|c|c|c}
    \hline
    \multirow{3}{*}{Method} & \multirow{3}{*}{Text} & \multicolumn{8}{c}{Metrics} \\
\cline{3-10}          &       & \multicolumn{2}{c|}{5\% labeled} & \multicolumn{2}{c|}{10\% labeled} & \multicolumn{2}{c|}{25\% labeled} & \multicolumn{2}{c}{50\% labeled} \\
\cline{3-10}          &       & mDice (\%) & mIoU (\%) & mDice (\%) & mIoU (\%) & mDice (\%) & mIoU (\%) & mDice (\%) & mIoU (\%) \\
    \hline
    MT \cite{ref5} & $\times$   & 74.25 & 64.14 & 75.54 & 64.22 & 77.11 & 67.65 & 79.31 & 69.05 \\
    BCP \cite{ref17} & $\times$    & 49.86 & 37.09 & 66.24 & 53.81 & 70.35 & 58.97 & 70.68 & 59.68 \\
    MC-Net \cite{ref18} & $\times$    & 70.57 & 59.89 & 71.30  & 59.54 & 73.34 & 62.27 & 75.09 & 64.18 \\
    SS-Net \cite{ref19} & $\times$    & 71.89 & 61.26 & 70.01 & 58.65 & 73.71 & 63.22 & 74.60  & 64.30 \\
    UCMT \cite{ref20} & $\times$    & 74.51 & 63.22 & 74.88 & 63.17 & 76.42 & 65.78 & 77.09 & 66.81 \\
    \hline
    CLIP \cite{ref9} & \checkmark   & 76.17 & 66.12 & 76.15 & 66.08 & 77.19 & 66.83 & 77.25 & 67.14 \\
    MedCLIP \cite{ref10} & \checkmark   & 76.47 & 66.84 & 76.72 & 67.05 & 77.01 & 66.59 & 78.30  & 67.99 \\
    MGCA \cite{ref11} & \checkmark   & 77.52 & 67.39 & 77.86  & 67.47 & 80.22 & \textbf{71.82} & 80.52 & 71.13 \\
    EGMA \cite{ref12} & \checkmark   & 77.69 & 68.21 & 78.54  & 67.82 & 79.43 & 68.56 & 81.04 & 71.51 \\
    DuSSS \cite{ref22} & \checkmark   & - & - & -  & - & 79.88 & 69.41 & 80.55 & 71.03 \\
    GraphCL \cite{ref23} & \checkmark   & 78.21 & 67.89 & 78.64  & 68.10 & 79.53 & 69.07 & 80.69 & 71.32 \\
    
    \hline
    Ours  & \checkmark & \textbf{80.44} & \textbf{70.63} & \textbf{81.67} & \textbf{70.43} & \textbf{82.21} & 71.36 & \textbf{82.93} & \textbf{73.55} \\
    \hline
    \end{tabular}}
  \label{tab1}%
\end{table*}%

\begin{figure*}[t]
\centering
\includegraphics[width=0.9\linewidth]{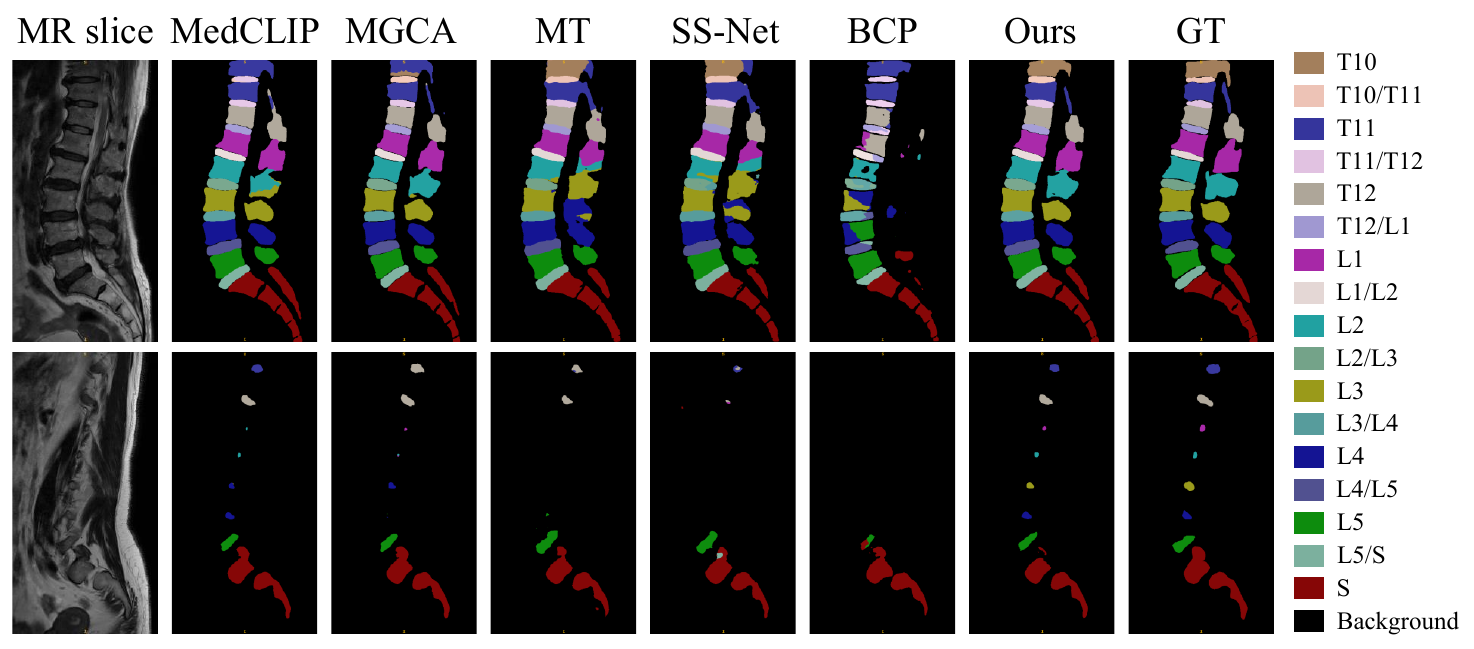}
\caption{Our method shows superior segmentation visualization results on 5\% labeled data, compared to SSL and VLM methods.}
\label{fig3}
\end{figure*}

\section{Experiments}
\textbf{Dataset.} The proposed method was evaluated on MRSpineSeg dataset \cite{ref14}. It contains T2-weighted MR volumetric images of 215 subjects. For each subject, the number of slices ranges from 12 to 18. There are 20 categories including 19 spinal structures and the background in the dataset. Not all images contain 19 spinal structures. The in-plane resolutions range from 512 × 512 to 1024 × 1024.

\begin{figure*}[t]
\centering
\includegraphics[width=0.9\linewidth]{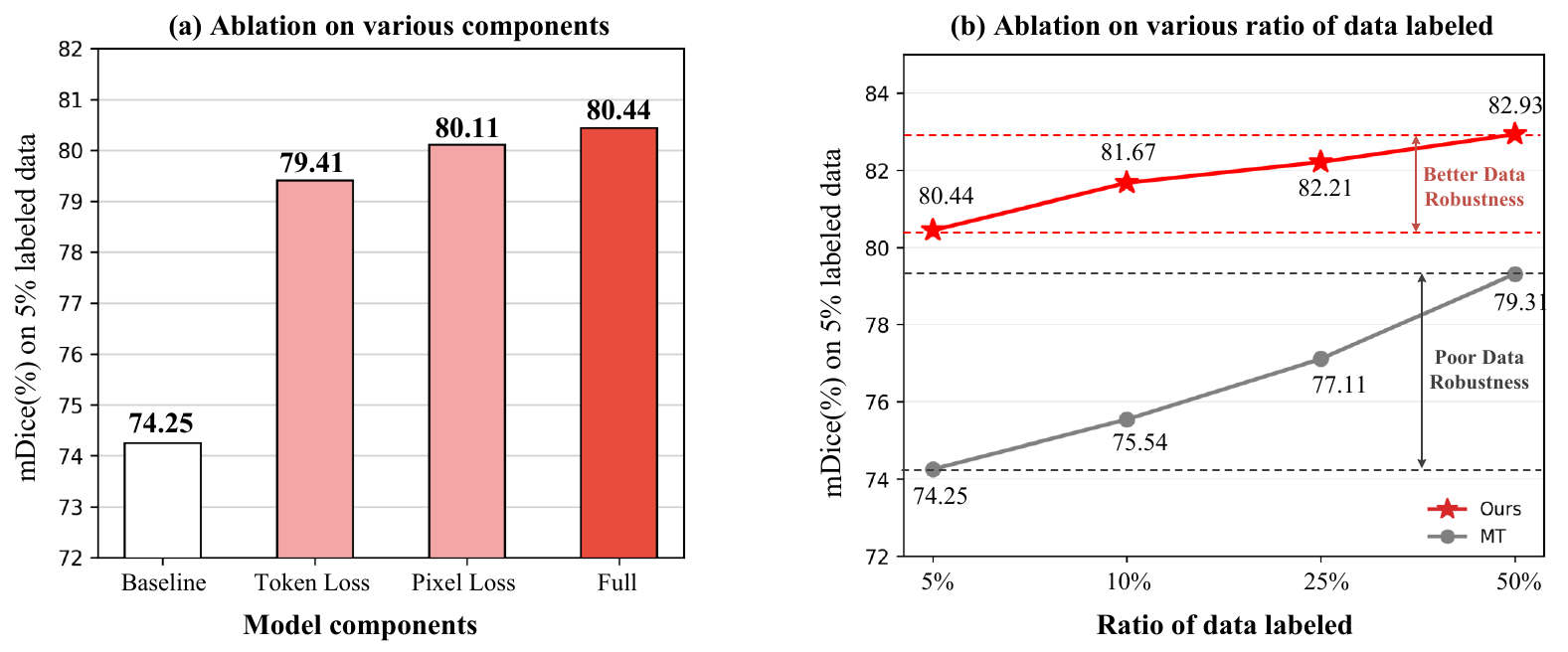}
\caption{(a) Ablation study on various components demonstrates the effectiveness of each component in CPS$^4$. (b) Ablation on different ratio of data labeled shows that CPS$^4$ has better data robustness with a subtle decrease when the labeled training data reduce to 5\%.}
\label{fig4}
\end{figure*}

\begin{figure*}[t]
\centering
\includegraphics[width=0.9\linewidth]{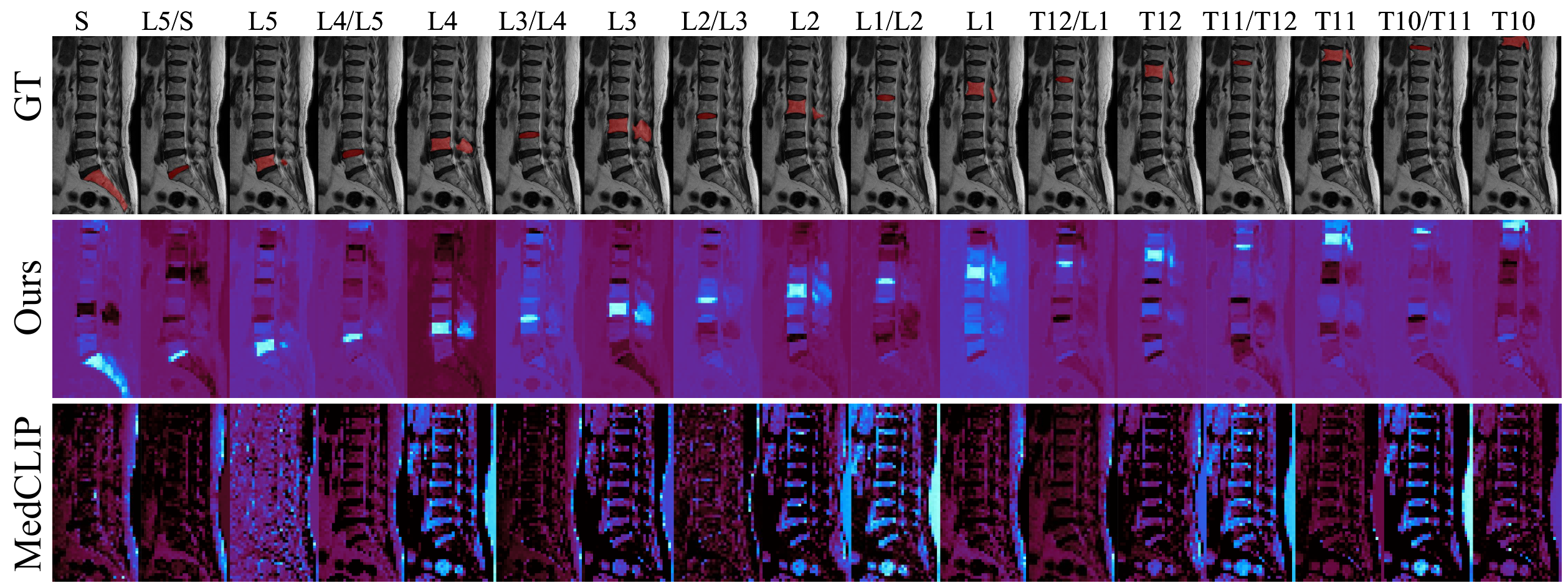}
\caption{Visualization of class-specific attention maps. The first row shows ground-truth annotations, the second row presents attention maps produced by our method, and the third row shows results from MedCLIP. Our method exhibits stronger consistency between class prompts and spine units, with attention highly aligned to each corresponding spine units.}
\label{fig5}
\end{figure*}

\vspace{4pt}
\noindent
\textbf{Implementation Details.} Our method is implemented by Pytorch. The operating system is Ubuntu 20.04.4 LTS with 24GB V100 GPU. For stage 1, the pretraining dataset is the labeled data from Stage 2, and each labeled MRI has a corresponding number of class prompts. We use Vit \cite{ref15} as vision encoder and BioClinicalBERT \cite{ref16} as text encoder. Adam optimizer with the learning rate of 1e-5 and ReduceLR scheduler are applied. The class prompt template is: A magnetic resonance image of a \{class name\}. For stage 2, in the class prompt guided segmentation map generation, each unlabeled MRI generates 20 attention maps corresponding to 20 classes, i.e., all classes in the dataset. Therefore, we do not need to know which classes each unlabeled image belongs to. The teacher-student network architecture is 2D ResUNet. The segmentation network is trained for 300 epochs using Adam optimizer with a weight decay of 1e-4 and a batch size of 8. The learning rate is set to 1e-4 initially and is lowered by 5 times at epoch 100 and 200.

\vspace{4pt}
\noindent
\textbf{Comparison Experiments.} We compare our CPS$^4$ with existing VLM and SSL methods, including MT \cite{ref5}, BCP \cite{ref17}, MC-Net \cite{ref18}, SS-Net \cite{ref19}, UCMT \cite{ref20}, CLIP \cite{ref9}, MedCLIP \cite{ref10}, MGCA \cite{ref11}, EGMA \cite{ref12}, DuSSS \cite{ref22}, and GraphCL \cite{ref23}. Fig. \ref{fig3} shows qualitative spine segmentation performance, where our method accurately localizes target regions, even in small object circumstances. MT, SS-Net, and BCP without class prompts all have more severe mis-segmentation, which indicates that the introduction of class prompts-guided segmentation map can help to produce more accurate segmentation. For quantitative experimental results (\autoref{tab1}), our method consistently outperforms all competing approaches across different labeling ratios in terms of both mDice and mIoU. Notably, under extremely limited supervision (5\% and 10\% labeled data), compared to suboptimal GraphCL with text prompts, CPS$^4$ improves mDice and mIoU by 2.23\% and 2.42\% on 5\% labeled data, and 3.03\% and 2.33\% on 10\% labeled data, demonstrating the superiority of CPS$^4$ in semi-supervised spine segmentation.

\vspace{4pt}
\noindent
\textbf{Ablation Study.} Ablation studies on diverse components and ratio of data labeled demonstrate the effectiveness of each component and powerful data robustness. Fig. \ref{fig4} (a) presents an ablation study on the key components of our method under the 5\% labeled data setting. Starting from the baseline model, incorporating the token loss $\mathcal{L}_{\mathrm{token}}$ leads to a substantial performance gain, and introducing the pixel loss $\mathcal{L}_{\mathrm{pixel}}$ also yields a large improvement, demonstrating the effectiveness of enhancing consistency between class prompts and spine units. When all components are combined, the performance reaches the highest mDice score, showing that each component contributes complementary benefits. Fig. \ref{fig4} (b) analyzes model robustness with respect to different ratios of labeled data. As the amount of labeled data increases from 5\% to 50\%, our method consistently outperforms the MT baseline by a clear margin. Notably, our model exhibits more stable performance gains across varying data ratios compared to MT.

\vspace{4pt}
\noindent
\textbf{Attention visualization Among Diverse Class Prompts.} Fig. \ref{fig5} shows qualitative visualization comparisons of class-specific attention maps. As observed, our method generates attention maps that are well aligned with the regions specified by the corresponding class prompts. For each class prompt, the highlighted regions accurately focus on the target spine unit, with less activation on irrelevant structures. This demonstrates strong consistency between class prompts and their corresponding spine units. In contrast, the attention maps generated by MedCLIP exhibit scattered activations. Such attention patterns indicate weak class-specific semantic binding, making it difficult to reliably associate the given class prompt with its spine unit. It shows that our method enables more reliable class prompt-driven pseudo-label generation, benefiting semi-supervised spine segmentation.

\section{Conclusion}
In this work, we propose CPS$^4$, a class prompt-drive semi-supervised spine segmentation framework consisting of two training stages. First, the class-specific consistency constrained VLM pretraining stage introduces token- and pixel-level attention losses to enforce the consistency between class prompts and corresponding spine units. Second, the class prompt-driven semi-supervised segmentation stage leverages the pretrained vision–text encoder to generate each class-specific binary segmentation map of the unlabeled image to guide the semi-supervised training. Extensive experiments demonstrate that CPS$^4$ achieves superior segmentation performance under various ratios of data labeled.

\section*{Acknowledgments.} This work was partly supported by Taishan Scholars Program of Shandong Province,  the National Natural Science Foundation of China (Grant No. 62173212 and 82220108007), and Shandong Province"Double Hundred Talent Plan”on 100 Foreign Experts and 100 Foreign Expert Teams (Grant No.WSR2023049).

\section*{Disclosure of Interests.} The authors have no competing interests to declare that are relevant to the content of this article.

\end{document}